\documentclass[journal]{IEEEtran}

\usepackage[utf8]{inputenc} %
\usepackage{hyperref}       %
\usepackage{url}            %
\usepackage{booktabs}       %
\usepackage{amsfonts}       %
\usepackage{amsmath}
\usepackage{nicefrac}       %
\usepackage{microtype}      %
\usepackage{paralist}
\usepackage{graphicx} %
\graphicspath{{img/}} %

\usepackage{pgfplots}
\usepgfplotslibrary{dateplot}
\usepackage{tikz}
\usetikzlibrary{positioning, shapes, fit, arrows}
\usetikzlibrary{external}

\newcommand{\citep}[1]{\cite{#1}}
\newcommand{\citet}[1]{\cite{#1}}
\newcommand{\argmin}{\text{argmin}}
\newcommand{\probe}[1]{$p_\text{#1}$}

\usepackage{color, colortbl}
\definecolor{LightCyan}{rgb}{0.8,1,1}

\newcommand{\diffrowcolor}[1]{}
\newcommand{\diffcolor}[1]{}

\begin{document}
\title{Unsupervised speech representation learning\\using WaveNet autoencoders}

\author{Jan~Chorowski, %
  Ron J. Weiss,  Samy Bengio, A\"{a}ron van den Oord
  \thanks{J. Chorowski is with the Institute of Computer Science,
    University of Wrocław, Poland e-mail: jan.chorowski@cs.uni.wroc.pl.}%
  \thanks{R. Weiss and S. Bengio are with Google Research. A. van den Oord is with DeepMind
    email: \{ronw, bengio, avdnoord\}@google.com.}%
}

\markboth{}%
{Chorowski \MakeLowercase{\textit{et al.}}: Unsupervised speech representation learning using WaveNet autoencoders}

\maketitle

\begin{abstract}
  We consider the task of unsupervised extraction of meaningful latent
  representations of speech by applying autoencoding neural networks
  to speech waveforms.
  The goal is to learn a representation able to capture high
  level semantic content from the signal, e.g.\ phoneme identities,
  while being invariant to confounding low level details in the signal
  such as the underlying pitch contour or background noise.
  {\diffcolor{blue}Since the learned representation is tuned to
    contain only phonetic content, we resort to using a high capacity
    WaveNet decoder to infer information discarded by the encoder from
    previous samples.
  Moreover, the} %
  behavior of autoencoder models depends on the kind of
  constraint that is applied to the latent representation. We compare
  three variants: a simple dimensionality reduction bottleneck, a Gaussian
  Variational Autoencoder (VAE), and a discrete Vector Quantized VAE
  (VQ-VAE). We analyze the quality of learned representations in terms of
  speaker independence, the ability to predict phonetic content, and the ability to
  accurately reconstruct individual spectrogram frames.  Moreover, for
  discrete encodings extracted using the VQ-VAE, we measure the
  ease of mapping them to phonemes. We introduce a regularization
  scheme that forces the representations to focus on the
  phonetic content of the utterance and report performance comparable
  with the top entries in the ZeroSpeech 2017 unsupervised acoustic
  unit discovery task.
\end{abstract}

\IEEEpeerreviewmaketitle

\section{Introduction}
Creating good data representations is important. The deep learning revolution
was triggered by the development of hierarchical representation learning
algorithms, such as stacked Restricted Boltzman Machines
\citep{hinton2006reducing} and Denoising Autoencoders
\citep{vincent2008extracting}. However, recent breakthroughs in computer vision
\citep{krizhevsky2012imagenet,szegedy2015going}, machine translation
\citep{bahdanau2014neural,wu2016google}, speech recognition
\citep{graves2013speech,chiu2017state}, and language understanding
\citep{wang2017gated,yu2018qanet} rely on large labeled datasets and make little
to no use of unsupervised representation learning. This has two drawbacks:
first, the requirement of large human labeled datasets often makes the
development of deep
learning models expensive.
Second, while a deep model may excel at solving a given task,
it yields limited insights into the problem domain, with main intuitions
typically consisting of visualizations of salient input patterns
\citep{zeiler2014visualizing,sundararajan2017axiomatic}, a strategy that is
applicable only to problem domains that are easily solved by humans.

In this paper we focus on evaluating and improving unsupervised speech
representations.  Specifically, we focus on
representations that separate selected speaker
traits, specifically speaker gender and identity, from phonetic
content, properties which are consistent with internal representations
learned by speech recognizers \cite{nagamine2017understanding,chorowski2018styletransfer}.
Such representations are desired in several tasks, such as low
resource automatic speech recognition (ASR), where only a small amount
of labeled training data is available.  %
In such scenario, limited amounts of data may be sufficient to
learn an acoustic model on the representation discovered without supervision, but
insufficient to learn the acoustic model and a data representation in a
fully supervised manner \cite{swietojanski2012unsupervised,thomas2013deep}.  %

We focus on representations learned with autoencoders applied
to raw waveforms and spectrogram features
and investigate the quality %
of learned representations on LibriSpeech
\cite{panayotov2015librispeech}. 
{\diffcolor{blue}We tune the learned latent representation to encode
  only phonetic content and remove other confounding
  detail. However, to enable signal reconstruction, we rely on an
  autoregressive WaveNet \cite{oord2016wavenet} decoder to infer
  information that was rejected by the encoder.
  The use of such a powerful decoder acts as an inductive bias,
  freeing up the encoder from using its capacity to represent
  low level detail and instead allowing it to focus on high level
  semantic features.}
We discover that best representations arise when ASR features,
such as mel-frequency cepstral coefficients (MFCCs) are used as
inputs, while raw waveforms are used as decoder
targets.  {\diffcolor{blue}This forces the system to also
  learn to generate sample level detail which was removed during
  feature extraction}.
Furthermore, we
observe that the Vector Quantized Variational Autoencoder (VQ-VAE)
\cite{oord2017neural} yields the best separation between the acoustic
content and speaker information. We investigate the interpetability of
VQ-VAE tokens by mapping them to phonemes, demonstrate the impact of model
hyperparameters on interpretability and propose a new regularization
scheme which improves the degree to which the latent representation can be mapped to 
the phonetic content.  Finally, we demonstrate strong performance on
the ZeroSpeech 2017 acoustic unit discovery task
\cite{dunbar2017zero}, which measures how discriminative a
representation is to minimal phonetic changes within an utterance.

\section{Representation Learning with Neural Networks}\label{sec:repr_learning}
Neural networks are hierarchical information processing models that
are typically implemented using layers of computational units. %
Each layer can be interpreted as a feature extractor whose
outputs are passed to upstream units
\citep{rumelhart1986learning}. Especially in the visual domain,
features learned with neural networks have been shown to create a
hierarchy of visual atoms \cite{zeiler2014visualizing} that
match some properties of the visual cortex
\citep{lee2008sparse}.
Similarly, when applied to audio waveforms, neural networks
have been shown to learn auditory-like frequency decompositions on
music \cite{dieleman2014end} and speech
\cite{jaitly2011learning,tuske2014acoustic,palaz2015analysis,sainath15waveform_cldnn}
in their lower layers.

\subsection{Supervised feature learning}
Neural networks can learn useful data representations in both
supervised and unsupervised manners. In the supervised case, features
learned on large datasets are often directly useful in similar but data-poor tasks.
For instance, in the visual domain, features discovered on
ImageNet \citep{deng2009imagenet} are routinely used as input
representations in other computer vision tasks
\citep{yosinski2014transferable}. Similarly, the speech community has used
bottleneck features 
extracted from networks trained on phoneme prediction tasks %
\citep{vesely2012language,yu2011improved} as feature representations for speech recognition systems.
Likewise, in
natural language processing, universal text representations can be
extracted from networks trained for machine translation
\citep{mccann2017learned} or language inference
\citep{bowman2015large,conneau2017supervised}.

\subsection{Unsupervised feature learning}
In this paper we focus on unsupervised feature learning. Since
no training labels are available we investigate autoencoders, i.e.,\ networks which
are tasked with reconstructing their inputs. Autoencoders use an
encoding network to extract a latent representation, which is then passed
through a decoding network to recover the original
data. Ideally, the latent representation preserves the salient
features of the original data, while being easier to analyze and work
with, e.g.\ by disentangling different factors of variation in the data, and discarding spurious patterns (noise). %
These desirable qualities %
are typically
obtained through a judicious application of regularization
techniques and constraints or bottlenecks (we use the two terms interchangeably). The representation learned
by an autoencoder is thus subject to two competing forces. On the one
hand, it should provide the decoder with information necessary
for perfect reconstruction and thus capture in the latents as much
of the input data characteristics as possible. On the other hand, the
constraints force some information to be discarded, preventing the
latent representation from being trivial to invert, e.g.\ by exactly passing through the input.
Thus the bottleneck is
necessary to force the network to learn a non-trivial data
transformation.

Reducing the dimensionality of the latent representation can serve as
a basic constraint applied to the latent vectors, with the
autoencoder acting as a nonlinear variant of linear low-rank
data projections, such as PCA or SVD
\citep{bishop2006pattern_pca}. However, such representations may be difficult to
interpret because the reconstruction of an
input depends on all latent features \cite{lee1999learning}. In contrast,
dictionary learning techniques, such as sparse
\cite{olshausen1996emergence} and non-negative \cite{lee1999learning}
decompositions, express each input pattern using a combination of a
small number of selected features out of a larger pool, which
facilitates their interpretability. Discrete feature learning using vector quantization can be
seen as an extreme form of sparseness in which the reconstruction uses only one
element from the dictionary.

The Variational Autoencoder (VAE) \citep{kingma2013auto} proposes a
different interpretation of feature learning which follows a probabilistic framework. The autoencoding
network is derived from a latent-variable generative model. First, a
latent vector $z$ is sampled from a prior distribution $p(z)$
(typically a multidimensional normal distribution). Then the data sample $x$ is
generated using a deep \emph{decoder} neural network with parameters
$\theta$ that computes $p(x|z; \theta)$. However, computing the exact
posterior distribution $p(z|x)$ that is needed during maximum
likelihood training is difficult. Instead, the VAE introduces a
variational approximation to the posterior, $q(z|x; \phi)$, which is modeled using an
\emph{encoder} neural network with parameters $\phi$. Thus the VAE resembles
a traditional autoencoder, in which the encoder produces
\emph{distributions} over latent representations, rather than deterministic
 encodings, while the decoder is trained on samples from this
distribution. Encoding and decoding networks are trained jointly
to maximize a lower bound on the log-likelihood of data point
$x$ \citep{kingma2013auto,higgins2017beta}:
\begin{align}\label{eq:vaecost}
  J_{\mathit{VAE}}(\theta, \phi; x) =\,& 
  \mathbb{E}_{q(z|x;\phi)}\left[\log p(x|z; \theta)\right] - \nonumber \\
  & \beta\, D_{KL}\left(q(z|x;\phi) \,||\, p(z)\right).
\end{align}

We can interpret the two terms of Eq.~\eqref{eq:vaecost} as the
autoencoder's reconstruction cost augmented with a penalty term applied to the hidden
representation. In particular, the KL divergence expresses the amount of
information in nats which the latent representation carries about the
data sample. Thus, it acts as an information bottleneck
\citep{alemi2016deep} on the latent representation, where
$\beta$ controls the trade-off between reconstruction
quality and the representation simplicity. 

An alternative formulation of the VAE objective explicitly constrains the amount of
information contained in the latent representation
\cite{kingma2016improved}:
\begin{align}\label{eq:vaecostfreebit}
  J_{\mathit{VAE}}(\theta, \phi; x) =\,&
  \mathbb{E}_{q(z|x;\phi)}\left[\log p(x|z; \theta)\right] - \nonumber \\
  & \max\left(B, D_{KL}\left(q(z|x;\phi) \,||\, p(z)\right)\right),
\end{align}
where the constant $B$ corresponds to the 
amount of \emph{free information} in $q$, because the model
is only penalized if it transmits more than
$B$ nats %
over the prior in the distribution over the latents.
Please note that for convenience we will often refer to information
content using units of bits instead of nats.

A recently proposed modification of the VAE, called the Vector Quantized
VAE \cite{oord2017neural},
replaces the continuous and stochastic latent vectors with
deterministically quantized versions. The
VQ-VAE maintains a number of prototype vectors $\{e_i, i=1,\ldots,K\}$. During the forward
pass, representations produced by the encoder are replaced with their
closest prototypes. Formally, let $z_e(x)$ be the output of the encoder
prior to quantization. VQ-VAE finds the nearest
prototype $q(x) = \argmin_i \|z_e(x) - e_i\|_2^2$ and uses it as the
latent representation $z_q(x) = e_{q(x)}$ which is passed to the
decoder. 
When using the model in downstream tasks, the learned
  representation can therefore be treated either as a distributed
  representation in which each sample is represented by a continuous
  vector, or as a discrete representation in which each sample is
  represented by the prototype ID (also called the token ID).

During the backward pass, the gradient of the loss with respect to the
pre-quantized embedding is approximated using the straight-through
estimator \citet{bengio2013estimating},
i.e.,\ $\frac{\partial\mathcal{L}}{\partial
  z_e(x)}\approx\frac{\partial\mathcal{L}}{\partial z_q(x)
}$\footnote{In TensorFlow this can be conveniently implemented using
  $z_q(x) = z_e(x) + \text{stop\_gradient}(e_{q(x)} - z_e(x))$}. 
The prototypes are trained by extending the learning objective with
terms which optimize quantization. Prototypes are forced to
lie close to vectors which they replace with an auxiliary cost, 
dubbed the commitment loss, introduced to encourage the encoder to
produce vectors which lie close to prototypes. Without the commitment loss VQ-VAE
training can diverge by emitting representations with unbounded magnitude. Therefore, VQ-VAE is trained
using a sum of three loss terms: the negative log-likelihood of the
reconstruction, which uses the straight-through estimator to bring the
gradient from the decoder to the encoder, and two VQ-related terms: the distance from each prototype to its assigned
vectors and the commitment cost \citep{oord2017neural}:
\begin{align}\label{eq:vqvaecost}
  \mathcal{L} =\, &\log p\big(x \mid z_q(x)\big) \nonumber\\
        &+ \|\text{sg}\big(z_e(x)\big) - e_{q(x)} \|^2_2
         + \gamma \|z_e(x) - \text{sg}(e_{q(x)}) \|_2^2,
\end{align}
where $\text{sg}(\cdot)$ denotes the \emph{stop-gradient} operation which zeros the gradient
with respect to its argument during backward pass. 

The quantization within the VQ-VAE acts as an information bottleneck. The encoder can be interpreted as a
probabilistic model which puts all probability mass on the selected
discrete token (prototype ID). Assuming a uniform prior distribution over $K$ tokens,
the KL divergence is constant and equal to $\log
K$. Therefore, the KL term does not need to be included in the VQ-VAE training
criterion in Eq.~\eqref{eq:vqvaecost} and instead becomes a hyperparameter
tied to the size of the prototype inventory.

The VQ-VAE was qualitatively shown to learn a
representation which separated the phonetic content within an
utterance from the identity of the speaker \citep{oord2017neural}. Moreover the discovered
tokens could be mapped to phonemes in a limited setting.

\subsection{Autoencoders for sequential data}

Sequential data, such as speech or text, often contain local
dependencies that can be exploited by generative models. In fact, purely
autoregressive models of sequential data, which predict the next
observation based on recent history, are very successful. For text,
these correspond to $n$-gram models \citep{jurafsky2009speech} and convolutional neural
language models \citep{dauphin2016language,bai2018empirical}. Similarly, %
WaveNet~\cite{oord2016wavenet} is a state-of-the-art autoregressive
model of time-domain waveform samples for text-to-speech synthesis.

\begin{figure}[t]
  \centering
  \newcommand{\conv}[2][3]{$\text{Conv}_{#1}(#2)$}
\newcommand{\stridedconv}[1]{$\text{StridedConv}_4(#1)$}
\newcommand{\proj}[1]{$\text{ReLU}(#1)$}
\newcommand{\linear}[1]{$\text{Linear}(#1)$}

\scalebox{0.85}{
\begin{tikzpicture}[auto, font=\small, node distance=1em and 1em, >=latex]
  \pgfdeclarelayer{back}
  \pgfsetlayers{back,main}

  \tikzstyle{input} = [align=center]
  \tikzstyle{block} = [draw, align=center, rectangle]
  \tikzstyle{conv} = [draw, align=center, rectangle]
  \tikzstyle{point} = [inner sep=0pt, minimum size=0, node distance=1ex]
  \tikzstyle{sum} = [draw, circle, inner sep=0, node distance=1ex, anchor=center]
  \tikzstyle{residual} = [looseness=3, relative, out=80, in=100] %

  \tikzstyle{componentlabel} = [align=left, anchor=east, font=\small\sffamily, inner ysep=0]
  \tikzstyle{shapelabel} = [align=left, anchor=south, right, inner xsep=0.7em, font=\scriptsize, scale=0.95, color=gray]

  \node [input, name=waveform] {waveform};
  \node [block, above=1.3cm of waveform] (featext) {MFCC + d + a \\feature extraction};
    \draw [->] (waveform) -- node[shapelabel, pos=0.75] {1D 16kHz} (featext);

  \node [conv, above=0.4cm of featext] (c1) {\conv{768}};
    \node [point, above=of c1] (c1s) {};
    \draw [->] (featext) -- node[shapelabel, name=conv_shape] {39D 100Hz} (c1);

  \node [conv, above=0.25cm of c1s] (c2) {\conv{768}};
    \node [sum, above=0.25cm of c2] (c2s) {+};
    \draw [-] (c1) -- (c1s);
    \draw [->] (c1s) -- node[shapelabel] {\\768D 100Hz} (c2);
    \draw [->] (c1s) edge[residual] (c2s);
    
  \node [conv, above=0.25cm of c2s] (c3) {\stridedconv{768}\\(stride = 2)};
    \node [point, above=of c3] (c3s) {};
    \draw [->] (c2) -- (c2s);
    \draw [->] (c2s) -- (c3); %

  \node [conv, above=0.25cm of c3s] (c4) {\conv{768}};
    \node [sum, above=0.25cm of c4] (c4s) {+};
    \draw [-] (c3) -- (c3s);
    \draw [->] (c3s) -- node[shapelabel] {\\768D 50Hz} (c4);

  \node [conv, above=0.25cm of c4s] (c5) {\conv{768}};
    \node [sum, above=0.25cm of c5] (c5s) {+};
    \draw [->] (c4) -- (c4s);
    \draw [->] (c4s) -- (c5); %
    \draw [->] (c3s) edge[residual] (c4s);

  \node [conv, above=0.25cm of c5s] (c6) {\proj{768}};
    \node [sum, above=0.25cm of c6] (c6s) {+};
    \draw [->] (c5) -- (c5s);
    \draw [->] (c5s) -- (c6); %
    \draw [->] (c4s) edge[residual] (c5s);
    
  \node [conv, above=0.25cm of c6s] (c7) {\proj{768}};
    \node [sum, above=0.25cm of c7] (c7s) {+};
    \draw [->] (c6) -- (c6s);
    \draw [->] (c6s) -- (c7);5 %
    \draw [->] (c5s) edge[residual] (c6s);

  \node [conv, above=0.25cm of c7s] (c8) {\proj{768}};
    \node [sum, above=0.25cm of c8] (c8s) {+};
    \draw [->] (c7) -- (c7s);
    \draw [->] (c7s) -- (c8); %
    \draw [->] (c6s) edge[residual] (c7s);

  \node [conv, above=0.25cm of c8s] (c9) {\proj{768}};
    \node [sum, above=0.25cm of c9] (c9s) {+};
    \draw [-] (c8) -- (c8s);
    \draw [->] (c8s) -- (c9); %
    \draw [->] (c7s) edge[residual] (c8s);
    \draw [->] (c8s) edge[residual] (c9s);
    \draw [->] (c9) -- (c9s);

  \node [componentlabel, above left=0.85cm and -0.2cm of c9, anchor=center] (enclabel) {Encoder};
  \begin{pgfonlayer}{back}
    \node [fill=%
      green!15, fit=(featext)(enclabel)(c3)(c9), inner xsep=0.2cm, xshift=0.1cm] {};
  \end{pgfonlayer}

  \tikzstyle{probe} = [draw, fill, circle, anchor=center, minimum size=0.05cm, inner sep=1pt] %
  \tikzstyle{probelabel} = [node distance=0]
  
  \node [point, right=1.65cm of c9s] (bn_in) {};
    \draw [-] (c9s) |- (bn_in)
          node [probe, pos=0.8] (p_encoded) {};
    \node [probelabel, above=of p_encoded] {\probe{enc}};

  \node [conv, right=0.4cm of bn_in] (proj) {\linear{64}};
    \draw [->] (bn_in) |- (proj);
  \node [conv, right=1.5cm of proj] (vq) {VQ};
    \draw [->] (proj) -- (vq)
          node [probe, pos=0.5] (p_prequant) {};
    \node [probelabel, above=of p_prequant] {\probe{proj}};
    \node[shapelabel, below=0pt of p_prequant] {64D 50Hz};

  \node [componentlabel, above=0.3cm of proj, anchor=center] (vqvae_label) {VQ-VAE\hspace{2.5ex}};

  \node [below right=1.05cm and 0.7cm of vqvae_label, anchor=east, font=\footnotesize\sffamily] (or_label) {or};

  \node [conv, below=1.05cm of proj] (vae_proj) {\linear{128}};
    \draw [->,dashed] (bn_in) |- (vae_proj);
  \node [conv, anchor=center, above right=-0.8em and 0.3cm of vae_proj] (split_mean) {$\mu$};
  \node [conv, anchor=center, below right=-0.82em and 0.3cm of vae_proj] (split_var) {$\sigma$};
  \node [point, right=0.35cm of vae_proj] (split) {};
    \draw [->] (vae_proj) -- (split);
  \node [conv, right=1.1cm of split] (sampling) {sample};
    \draw [->] (split_mean) -- (sampling)
          node [probe, pos=0.5] (vae_p_encoded) {};
    \node [probelabel, above=of vae_p_encoded] {\probe{proj}};
    \draw [->] (split_var) -- (sampling);
  \node [componentlabel, above=0.3cm of vae_proj, anchor=east] (vae_label) {VAE\hspace{0.5ex}};
  \begin{pgfonlayer}{back}
    \node [fill=red!10, fit=(vae_label)(vae_proj)(split_mean)(split_var)(sampling), inner xsep=0.2cm] {};
  \end{pgfonlayer}

  \node [above right=2cm and 0cm of sampling, anchor=east] (vqvae_left) {};
  \begin{pgfonlayer}{back}V
    \node [fill=red!15, fit=(vqvae_label)(proj)(vq)(vqvae_left), inner xsep=0.2cm] {};
  \end{pgfonlayer}

  \node [below=1.5cm of or_label, anchor=center, font=\footnotesize\sffamily] {or};

  \node [conv, below=1.15cm of vae_proj] (ae_proj) {\linear{64}};
    \draw [->,dashed] (bn_in) |- (ae_proj);
  \node [componentlabel, above=0.3cm of ae_proj, anchor=east] (ae_label) {AE\hspace{1.5ex}};
  \node [below right=1cm and 0cm of sampling, anchor=east] (ae_left) {};
  \begin{pgfonlayer}{back}
    \node [fill=red!10, fit=(ae_label)(ae_proj)(ae_left), inner xsep=0.2cm] {};
  \end{pgfonlayer}

  \node [point, right=3cm of ae_proj] (bn_out) {};
    \draw [-] (vq) -| (bn_out);
    \draw [-,dashed] (sampling) -| (bn_out);
    \draw [-,dashed] (ae_proj) -- (bn_out);

  \node [conv, below left=0.9cm and 0.52cm of bn_out] (jitter) {$\text{jitter}(0.12)$};
    \draw [->] (bn_out) |- (jitter)
          node [probe, pos=0.25] (p_sample) {};
    \node [probelabel, left=of p_sample] {\probe{bn}};
  \node [conv, below=0.22cm of jitter] (conditioning) {\conv[3]{128}};
    \draw [->] (jitter) -- (conditioning) {};
    \node [conv, below=0.5cm of conditioning] (upsample) {upsample};
    \draw [->] (conditioning) -- (upsample)
          node [probe, pos=0.45] (p_conditioning) {};
    \node [probelabel, left=of p_conditioning] {\probe{cond}};
    \node [shapelabel, right=-0.5ex of p_conditioning, name=cond_shape] {128D 50Hz};

  \node [input, name=speakerid, right=5.35cm of waveform, align=center] {speaker\\one-hot};
  \node [conv, below=0.4cm of upsample] (concat) {concat};
    \draw [->] (upsample) --  node[shapelabel, pos=0.5] {128D 16kHz} (concat);
    \draw [->] (speakerid) |- node [shapelabel, left, pos=0.017] {$N_s$} (concat);

  \tikzset{
    wavenet/.pic={code={
      \tikzset{scale=1/4}
      \foreach \x in {1,...,8} \draw[fill] (\x, 0) circle (0.1cm);
      \foreach \x in {2,4,6,8} \draw (\x, 0) -- (\x, 1) -- (\x-1, 0);
      \foreach \x in {2,4,6,8} \draw[fill] (\x, 1) circle (0.1cm);
      \foreach \x in {4,8} \draw (\x, 1) -- (\x, 2) -- (\x-2, 1);
     \foreach \x in {4,8} \draw[fill] (\x, 2) circle (0.1cm);
      \draw (8, 2) -- (8, 3) -- (4, 2);
      \draw[fill] (8, 3) circle (0.1cm);
      \node[font=\small, align=left, anchor=north west] at (0.1, 6.0) {WaveNet cycle\\(10 layers)};
      \draw (8.7, -0.8) rectangle (0, 6.0);
    }}
  }
  \pic [local bounding box=wn1, anchor=center, below left=0.9cm and 2.4cm of concat] {wavenet};
    \node[point, above=0.1cm of wn1] (wn1_in) {};
    \draw [->] (wn1_in) -- (wn1);
    \draw [->] (concat) |- (wn1);
 \node[point, below=0.15cm of wn1] (wn1_out) {};
  \pic [local bounding box=wn2, below left=3.1cm and 2.4cm of concat] {wavenet};
    \draw [->] (concat) |- node[shapelabel, pos=0.08, xshift=-0.7ex] {128$\,+N_s$\\16kHz} (wn2);
    \draw [->] (wn1) -- node[shapelabel] {256D 16kHz} (wn2);

  \node [sum, below=0.3cm of wn2] (wn2s) {+};
    \draw [->] (wn2) -- (wn2s);
    \draw [->] (wn1_out) edge[residual, looseness=1.9, out=-100, in=-90] (wn2s);
  \node [conv, right=0.3cm of wn2s] (wn_out1) {\proj{256}};
    \draw [->] (wn2s) -- (wn_out1);
  \node [conv, below=0.22cm of wn_out1] (wn_out2) {\proj{256}};
    \draw [->] (wn_out1) -- (wn_out2);
  \node [conv, below=0.22cm of wn_out2] (wn_softmax) {softmax};
    \draw [->] (wn_out2) -- (wn_softmax);
  \node [conv, left=0.4cm of wn_softmax] (wn_sample) {sample};
    \draw [->] (wn_softmax) -- (wn_sample);
    \node[point, left=0.7cm of wn_sample] (wn_out) {};

  \draw [-] (wn_sample) -- (wn_out); %
  \draw [->] (wn_out) |- (waveform);
  \node [point, above=6.3cm of wn_out] (wn_ar) {};
  \draw [-] (wn_out) -- (wn_ar);
  \draw [-] (wn_ar) -| (wn1_in);

  \node [componentlabel, left=1.15cm of jitter] (declabel) {Decoder};
  \begin{pgfonlayer}{back}
    \node [fill=blue!15, fit=(jitter)(conditioning)(upsample)(declabel)(wn1)(wn_sample)(cond_shape), inner ysep=0.1cm, inner xsep=0.08cm, xshift=-0.08cm] {};
  \end{pgfonlayer}

\end{tikzpicture}
}
  \vspace{-3ex}
  \caption{The proposed model is conceptually divided into 3 parts: an
    \emph{encoder} (green), made of a residual convnet that computes a
    stream of latent vectors (typically every 10ms or 20ms) from a time-domain
    waveform sampled at 16~kHz, which are passed through a \emph{bottleneck}
    (red) before being used to condition a WaveNet \emph{decoder} (blue) which
    reconstructs the waveform using two additional information streams: an
    autoregressive stream which predicts the next sample based on past samples,
    and global conditioning which represents the identity of the input speaker
    (one out of $N_s$ total training speakers).
    We experiment with three bottleneck variants: a simple dimensionality
    reduction (AE), a sampling layer with an additional Kullback-Leibler penalty
    term (VAE), or a discretization layer (VQ-VAE).
    Intuitively, this bottleneck encourages the encoder to discard portions of
    the latent representation which the decoder can infer from the two other
    information streams.
    For all layers, numbers in parentheses indicate the number of output
    channels, and subscripts denote the filter length.  Locations of ``probe''
    points which are used in Section~\ref{sec:experiments} to evaluate the
    quality of the learned representation are denoted with black dots.}
  \label{fig:experimental_system}
\end{figure}
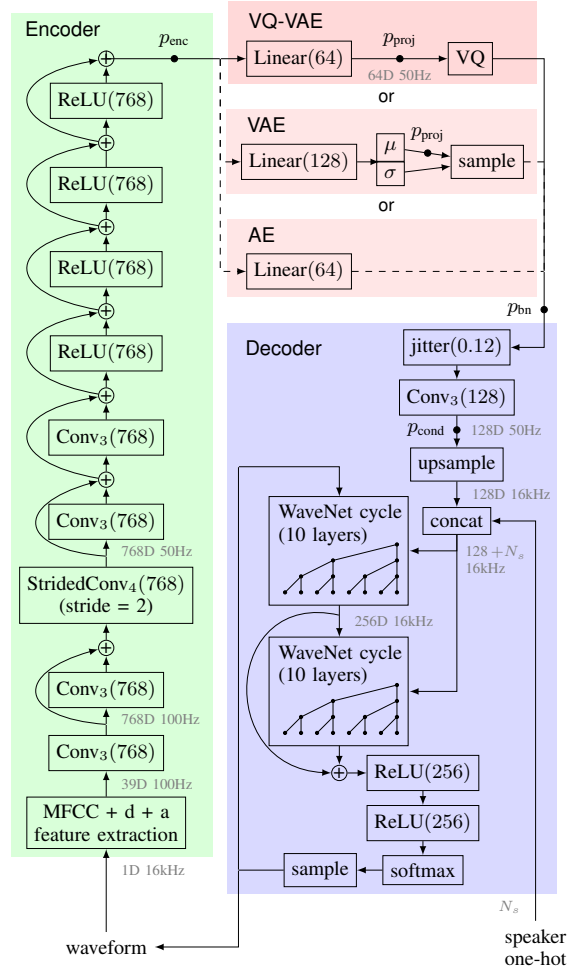

A downside of such autoregressive models is that they do not explicitly
produce latent representations of the data. However, it is possible to
combine an autoregressive sequence generation model with an encoder
tasked with extraction of latent representations. Depending on the
use case, the encoder can process the whole utterance, emit a single
latent vector and feed it to an autoregressive decoder
\cite{bowman2015large,hsu2018hierarchical} or the encoder can periodically emit vectors
of latent features to be consumed by the decoder
\cite{oord2017neural,gulrajani2016pixelvae}. We concentrate on
the latter solution.

Training mixed latent variable and autoregressive models is prone to
latent space collapse, in which the decoder learns to ignore the
constrained latent representations and only uses the unconstrained
signal coming through the autoregressive 
path. For the VAE, this collapse can be prevented by annealing the
weight of the KL term and using the free-information formulation
in Eq.~\eqref{eq:vaecostfreebit}. The VQ-VAE is naturally resilient to the
latent collapse because the KL term is a hyperparameter which is
not optimized using gradient training of a given model. We defer
further discussion of this topic to Section \ref{sec:rel_work}.

\section{Model Description}
The architecture of our model is presented in
Figure~\ref{fig:experimental_system}. The encoder reads a sequence of
either raw audio samples, or of audio features\footnote{To keep the
  autoencoder viewpoint, the feature extractor can be interpreted as
  a fixed signal processing layer in the encoder.} and extracts a
sequence of hidden vectors, which are passed through a bottleneck to
become a sequence of latent representations. The frequency at which
the latent vectors are extracted is governed by the number of strided
convolutions applied by the encoder.

The decoder reconstructs the utterance by conditioning a WaveNet
\cite{oord2016wavenet} network on
the latent representation extracted by the encoder and, separately, on
a speaker embedding.
Explicitly conditioning the decoder on speaker identity frees the
encoder from having to capture speaker-dependent information in the
latent representation.
Specifically, the decoder 
\begin{inparaenum}[(i)]
\item takes the encoder's output,
\item optionally applies a
stochastic regularization to the latent vectors (see Section~\ref{sec:jitter}),
\item then combines latent
vectors extracted at neighboring time steps using
convolutions and 
\item upsamples them to the output frequency. 
\end{inparaenum}
Waveform samples
are reconstructed with a WaveNet 
that combines all conditioning sources: autoregressive information
about \emph{past} samples, global information about the speaker,
and latent information about \emph{past and future}
samples extracted by the encoder. We find that the encoder's bottleneck and the
proposed regularization is crucial
in extracting nontrivial representations of data. With no bottleneck,
the model is prone to learn a simple reconstruction strategy which
makes verbatim copies of future samples. We also note that the encoder
is speaker independent and requires only speech data, while the decoder
also requires speaker information.

We consider three forms of bottleneck:
\begin{inparaenum}[(i)]
\item simple dimensionality reduction,
\item a Gaussian VAE with different latent representation dimensionalities and
  different capacities %
  following Eq.~\eqref{eq:vaecostfreebit}, and
\item a VQ-VAE with different number of quantization prototypes.
\end{inparaenum}
All bottlenecks are
optionally followed by the dropout inspired time-jitter regularization
described below. Furthermore, we experiment with different input and output
representations, using raw waveforms, log-mel filterbank, and mel-frequency cepstral
coefficient (MFCC) features which discard pitch information present in the spectrogram.

\subsection{Time-jitter regularization}\label{sec:jitter}

We would like the model to learn a representation of speech which
corresponds to the slowly-changing phonetic content within an utterance: a
mostly constant signal that can abruptly change at phoneme boundaries.

Inspired by the slow features analysis \citep{wiskott2002slow} we
first experimented with penalizing time differences
between encoder representation either before or after the bottleneck. However, this
regularization resulted in a collapse of the latent space -- the model
learned to output a constant encoding. This is a common problem of sequential
VAEs that use loss terms to regularize the latent encoding
\citep{bowman2015generating}.

Reconsidering the problem we realized that we want each frame's
representation to correspond to a meaningful phonetic unit. Thus we
want to prevent the system from using consecutive latent vectors as
individual units. Put differently, we want to prevent latent vector
co-adaptation. We therefore introduce a dropout-inspired
\citep{srivastava2014dropout} time-jitter regularizer, also
reminiscent of Zoneout \citep{krueger2016zoneout} regularization for
recurrent networks.  During training, each latent vector can replace
either one or both of its neighbors. As in dropout, this prevents the
model from relying on consistency across groups of
tokens. Additionally, this regularization also promotes latent
representation stability over time: a latent vector extracted at time
step $t$ must strive to also be useful at time steps $t-1$ or
$t+1$. In fact, the regularization was crucial for reaching good
performance on ZeroSpeech at higher token extraction frequencies.

The regularization layer is inserted right after the encoder's
bottleneck (i.e., after dimensionality reduction for regular
autoencoder, after sampling a realization of the latent layer for the
VAE and after discretization for the VQ-VAE). It is only enabled during
training. For each time step we independently sample whether it is to
be replaced with the token right after or before it. We do not copy a
token more than one timestep.

\section{Experiments}\label{sec:experiments}

We evaluated models on two datasets: LibriSpeech
\citep{panayotov2015librispeech} (clean subset) and ZeroSpeech
2017 Contest Track 1 data \citep{dunbar2017zero}. Both datasets have
similar characteristics: multiple speakers, clean, read speech (sourced from audio
books) recorded at a sampling rate of 16~kHz. %
Moreover the ZeroSpeech challenge controls the amount of
per-speaker data with the majority of the data being uttered by only a
few speakers.

Initial experiments, presented in section \ref{sec:exp_disentag}, 
compare different bottleneck variants and establish what type of information
from the input audio is preserved in the continuous latent representations produced by
the model at the four different probe points pictured in
Figure~\ref{fig:experimental_system}.
Using the representation computed at each probe point, we measure performance on several prediction tasks:
phoneme prediction (per-frame accuracy),
speaker identity and gender prediction accuracy, and $L_2$ reconstruction error
of spectrogram frames.
We establish that the VQ-VAE learns
latent representations with strongest disentanglement between the
phonetic content and speaker identity, and focus on this
architecture in the following experiments.

In section \ref{sec:exp_interp} we analyze the interpretability of VQ-VAE
tokens by
mapping each discrete token to the most frequent corresponding phoneme in a forced
alignment of a small labeled data set (LibriSpeech dev) and report the
accuracy of the mapping on a separate set (LibriSpeech test).
Intuitively, this captures the interpretability of individual tokens.

We then
apply the VQ-VAE to the ZeroSpeech 2017 acoustic unit discovery task
\citep{dunbar2017zero} in section \ref{sec:exp_zs}.  This task evaluates how discriminative the
representation is with respect to the phonetic class. Finally, in section
\ref{sec:exp_hyper} we measure the impact of different hyperparameters on
performance.

\begin{figure*}[t]
  \centering
  \includegraphics[width=0.98\textwidth]{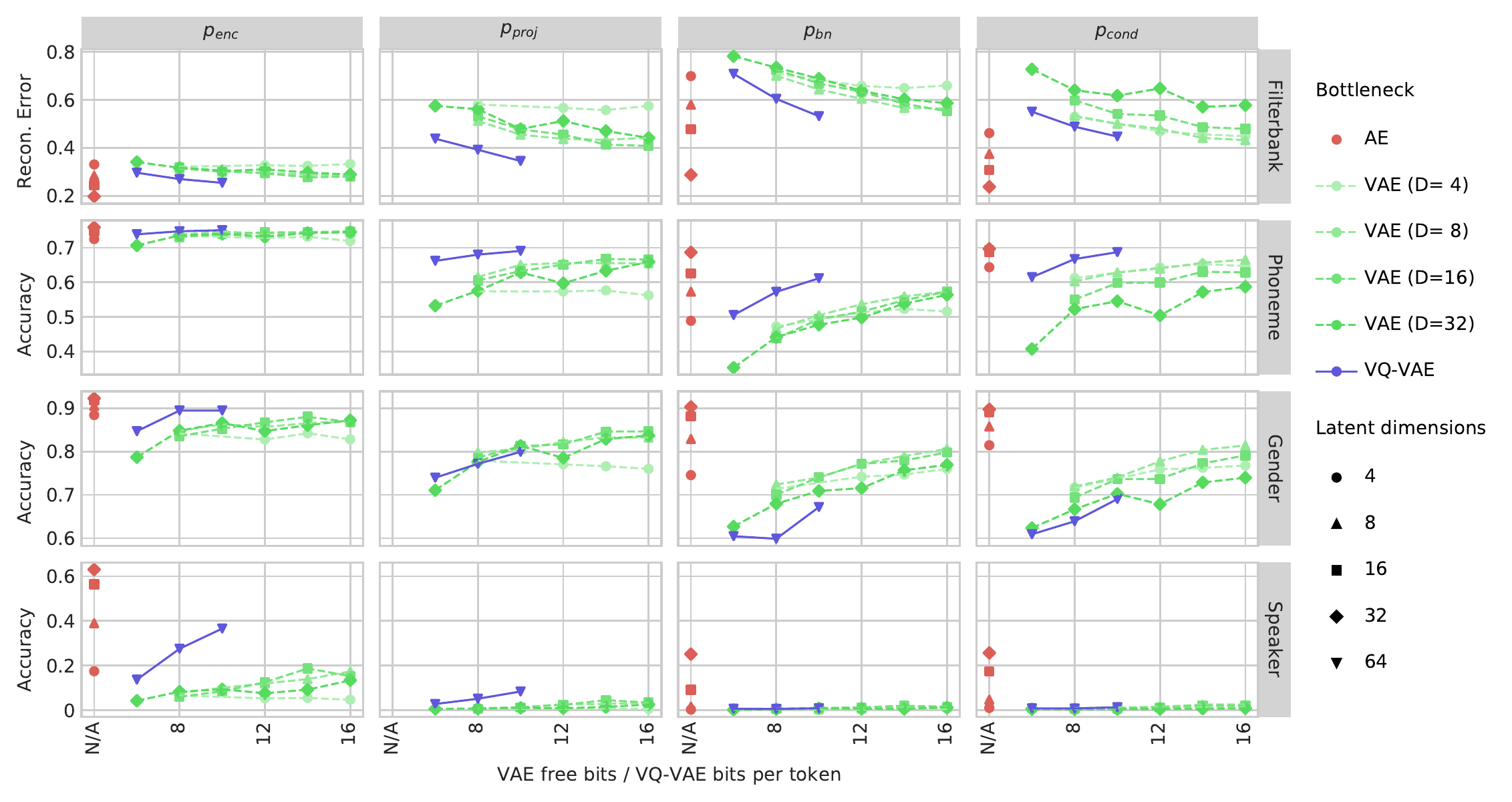}
  \vspace{-2ex}
  \caption{
    Accuracy of predicting signal characteristics
    at various probe locations in the network. Among the three
    bottlenecks evaluated, VQ-VAE discards the most speaker-related
    information at the bottleneck, while preserving the most phonetic
    information. For all bottlenecks, the representation coming out of
    the encoder yields over 70\% accurate framewise  phoneme
    predictions. Both the simple AE and VQ-VAE
    preserve this information in the bottleneck (the accuracy drops to
    50\%-60\% depending on the bottleneck's strength). However, the
    VQ-VAE discards almost all speaker information (speaker
    classification accuracy is close to 0\% and gender prediction
    close to 50\%). This causes the VQ-VAE representation to perform
    best on the acoustic unit discovery task -- the representation
    captures the phonetic content while being invariant to speaker identity.}
  \label{fig:main_probes}
\end{figure*}

\begin{figure}[t]
  \centering
  \includegraphics[width=\columnwidth, trim=0 1.5ex 0 0, clip]{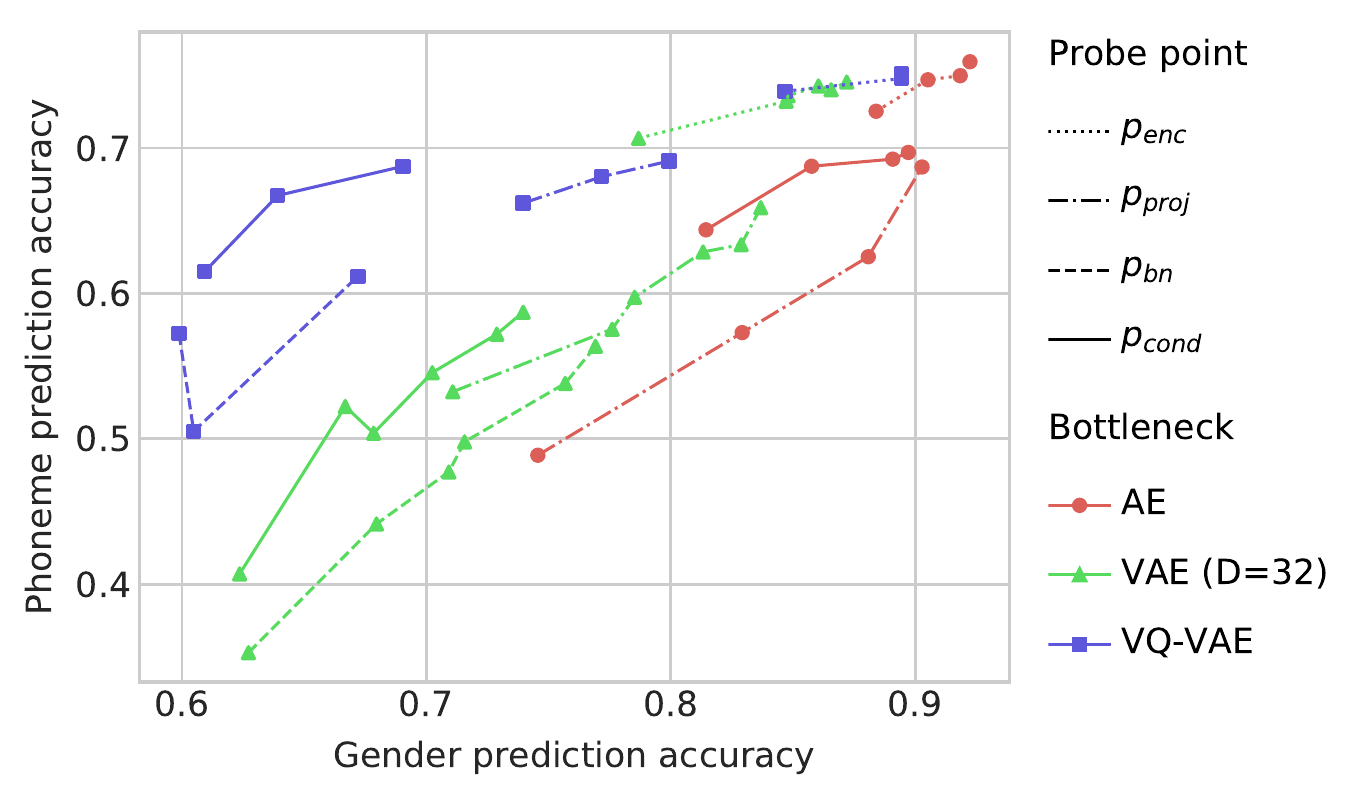}
  \caption{Comparison of gender and phoneme prediction accuracy for
    different bottleneck types and probe points. The decoder is
    conditioned on the speaker, thus the gender information can be
    recovered and the bottleneck should discard it. 
    While information is present at the $p_\text{enc}$ probe. The AE
    and VAE models tend to similarly discard both gender and phoneme
    information at other probe points. On the other hand, VQ-VAE selectively discards gender
    information.} 
  \label{fig:gender_vs_phone}
\end{figure}

\subsection{Default model hyperparameters}
Our best models used MFCCs as the encoder input, %
but reconstructed raw waveforms at the decoder output.
We used standard 13 MFCC features extracted every
10ms (i.e.,\ at a rate of 100~Hz) and augmented with their temporal first and second
derivatives. Such features were originally designed for speech
recognition and are mostly invariant to pitch and similar confounding detail
in the audio signal.  The encoder had 9 layers each
using 768 units with ReLU activation, organized into the following groups:
2 preprocessing convolution layers with filter length 3 and residual
connections,
1 strided convolution length reduction layer with filter length 4 and stride 2
(downsampling the signal by a factor of two),
followed by 2 convolutional layers with length 3 and residual connections,
and finally 4 feedforward ReLU layers with residual connections.
The resulting latent vectors were extracted at 50~Hz (i.e., every second frame), with each
latent vector depending on a receptive field of 16 input frames.
We also used an alternative encoder with two length reduction layers, 
which extracted latent representation at 25~Hz with a receptive field of 30 frames.

When unspecified, the latent representation was 64 dimensional and when applicable constrained
to 14 bits. Furthermore, for the VQ-VAE we used the recommended $\gamma=0.25$ \cite{oord2017neural}.

The decoder applied the randomized time-jitter
regularization (see Section \ref{sec:jitter}). During training each latent
vector was replaced with either of its neighbors with probability 0.12.
The jittered latent sequence was passed through a single convolutional layer
with filter length 3
and 128 hidden units to mix information across neighboring
timesteps. The representation was then upsampled 320 times (to
match the 16kHz audio sampling rate) and concatenated with a one-hot vector
representing the current speaker to form the conditioning input of an
autoregressive WaveNet \citep{oord2016wavenet}. The WaveNet was composed of 20
causal dilated convolution layers, each using 368 gated units with residual
connections, organized into two ``cycles'' of 10 layers with dilation
rates $1, 2, 4, \ldots, 2^{9}$.  The conditioning signal was passed
separately into each layer. The signal from each layer of the WaveNet
was passed to the output using skip-connections.  Finally, the signal
was passed through 2 ReLU layers with 256 units.
A Softmax was applied to compute the next sample probability.
We used 256 quantization levels after mu-law companding \citep{oord2016wavenet}.

All models were trained on minibatches of 64 sequences of length 5120
time-domain samples (320 ms)
sampled uniformly from the training dataset. Training
a single model on 4 Google Cloud TPUs (16 chips) took a week. We used the Adam
optimizer \citep{kingma2014adam} with initial learning rate
$4\times10^{-4}$ which %
was halved after 400k,
600k, and 800k steps.  Polyak averaging \citep{polyak1992acceleration}
was applied to all checkpoints used for model evaluation.

\begin{figure*}[t]
  \centering
  \includegraphics[width=0.9\textwidth, trim=0 2.4ex 0 2.1ex, clip]{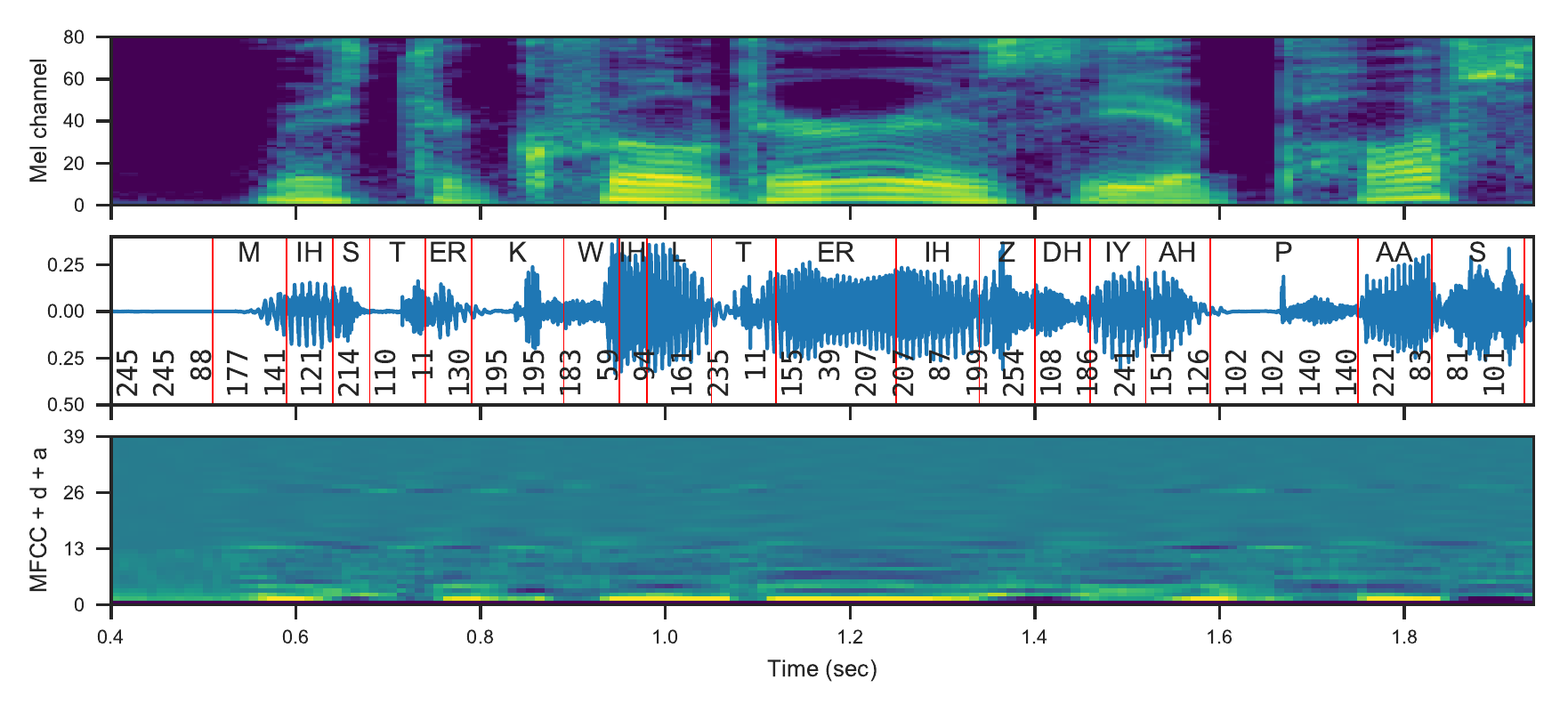}
  \caption{Example token sequence extracted by the proposed VQ-VAE model.
    Bottom: input MFCC features.
    Middle: Target waveform samples overlaid with extracted token IDs (bottom),
    and ground truth phoneme identities (top) with boundaries between plotted in
    red.
    Note the transient ``T'' phoneme at 0.75 and 1.1 seconds is consistently
    associated with token 11.
    The corresponding log-mel spectrogram is shown on the top.
  }
  \label{fig:example_signal}
\end{figure*}

\subsection{Bottleneck comparison}\label{sec:exp_disentag}

We train models on LibriSpeech and
analyze the information captured in the hidden representations surrounding
the autoencoder bottleneck at each of the four probe points shown in
Figure~\ref{fig:experimental_system}:
\begin{itemize}[]%
  \item \probe{enc} (768 dim) encoder output prior to the bottleneck,
  \item \probe{proj} (64 dim) within the bottleneck after projecting to lower
    dimension,
  \item \probe{bn} (64 dim) bottleneck output, corresponding to the quantized
    representation in VQ-VAE, or a random sample from the variational
    posterior in VAE, and
  \item \probe{cond} (128 dim) after passing \probe{bn} through a convolution
    layer which captures a larger receptive field over the latent encoding.
\end{itemize}
At each probe point, we train separate MLP networks with 2048 hidden units on each of four tasks:
classifying speaker gender and identity for the whole segment (after
average pooling latent vectors across the full signal),
predicting phoneme class at each frame (making several predictions per
latent vector\footnote{Ground truth phoneme labels and filterbank features have a frame rate of 100~Hz, while the latent representation is computed at a lower rate.}), and
reconstructing log-mel filterbank features\footnote{We
  also experimented with probes that reconstructed MFCCs, but the results were strongly correlated with
  those on filterbanks so we do not include them.
  We did not evaluate waveform reconstruction because training a full WaveNet for
  each probe point was too expensive.} in each frame (again predicting
several consecutive frames from each latent
vector).
A representation which captures the high level semantic content from the signal,
while being invariant to nuisance low-level signal details, will have a high
phoneme prediction accuracy, and high spectrogram reconstruction error.
A disentangled representation should additionally have low speaker prediction
accuracy, since this information is explicitly made available to the decoder
conditioning network, 
and therefore need not be preserved in the latent encoding \cite{moyer_invariant_2018}.
Since we are primarily interested in discovering what information is present in
the constructed representations we report the training
performance and do not tune probing networks for
generalization.

A comparison of models using each of the three bottlenecks with
different hyperparameters (latent dimensionality and
bottleneck bitrate) is %
presented in Figure~\ref{fig:main_probes},
illustrating the degree of information propagation
  through the network.
In addition, Figure~\ref{fig:gender_vs_phone},
  highlights the separation of phonetic content and speaker identity
  obtained using different configurations.

Figure~\ref{fig:main_probes} shows that each bottleneck type consistently
discards information between the \probe{enc} and \probe{bn} probe locations, as
evidenced by the reduced performance on each task.
The bottleneck also impacts information content in 
preceding layers. Especially for the vanilla autoencoder (AE),
which simply reduces dimensionality, the speaker prediction accuracy and filterbank
reconstruction loss at \probe{enc} depend on the width of the
bottleneck, with narrower widths causing more information to be
discarded in lower layers of the encoder. Likewise, VQ-VAEs and AEs
yielded better filterbank reconstructions and
speaker identity prediction at \probe{enc} compared to
VAEs with matching dimensionality and bitrate, which
  corresponds to the logarithm of the number of tokens for the VQ-VAE, and
  the KL divergence from the prior for the VAE, which we 
  control by setting the number of allowed free bits.

As expected, \emph{AE} discards the least
information.  At \probe{cond} the representation remains highly
predictive about both speaker and phonemes, and its filterbank
reconstructions are the best among all configurations.
However, from an unsupervised learning standpoint, the AE latent
representation is less %
useful because it mixes all properties of the source signal.

In contrast, \emph{VQ-VAE} models produce a representation which is highly
predictive of the phonetic content of the signal while effectively discarding
speaker identity and gender information.
At higher bitrates, phoneme prediction is about as accurate as for the AE. 
Filterbank reconstructions are also less accurate. We
observe that the speaker information is discarded primarily during
the quantization step between \probe{proj} and
\probe{bn}. Combining several latent vectors in the
\probe{cond} representation results in more accurate phoneme
predictions, but the additional context does not help to recover speaker
information. This phenomenon is highlighted %
in Figure~\ref{fig:gender_vs_phone}.
Note that VQ-VAE models showed little dependence on the bottleneck dimension,
so we present results at the default setting of 64.

Finally, \emph{VAE} models separate speaker and phonetic information better than
simple dimensionality reduction, but not as well as VQ-VAE. The VAE
discards phonetic and speaker information more uniformly than VQ-VAE:
at \probe{bn}, VAE's phoneme predictions are
less accurate, while its gender predictions are more accurate. Moreover,
combining information across a wider receptive field at \probe{cond} does not improve
phoneme recognition as much as in VQ-VAE models. The sensitivity to the 
bottleneck dimensionality, seen in
Figure~\ref{fig:main_probes} 
is also surprising, with narrower VAE bottlenecks discarding less
information than wider ones. This may be due to the stochastic
operation of the VAE: to provide the same KL divergence as at %
low bottleneck dimensions, more noise needs to be added at high
dimensions. This noise may mask information present in the
representation. 

Based on these results we conclude that the VQ-VAE bottleneck is most
appropriate for learning latent representations which capture phonetic content
while being invariant to the underlying speaker identity.

\subsection{VQ-VAE token interpretability} %
\label{sec:exp_interp}

Up to this point we have used the VQ-VAE as a
  bottleneck that quantizes latent vectors. In this section we 
  seek an interpretation of the discrete prototype IDs, evaluating whether VQ-VAE
  tokens can be mapped to phonemes, the underlying discrete constituents
  of speech sounds. Example token IDs are pictured in the middle pane of
  Figure~\ref{fig:example_signal}, where we can see that the token 11 is
  consistently associated with the transient ``T'' phone.
  To evaluate whether other tokens have
  similar interpretations,
we measured the
frame-wise phoneme recognition accuracy in which each token was mapped
to one out of 41 phonemes.
We used the 460 hour clean LibriSpeech training set for unsupervised
training, and used labels from the clean dev subset to associate each token with the most probable phoneme.
We evaluated the mapping by computing frame-wise phone recognition
accuracy on the clean test set at a frame rate of 100~Hz. The ground-truth phoneme boundaries were obtained
from forced alignments using the Kaldi \texttt{tri6b} model from the \texttt{s5} LibriSpeech recipe
\citep{povey2011kaldi}.

\begin{table}[t]
  \caption{LibriSpeech frame-wise phoneme recognition accuracy. VQ-VAE models consume
    MFCC features and extracted tokens at 25~Hz.}
  \label{tb:best_lbr}
  \centering
  \setlength{\tabcolsep}{1.25ex}
  \begin{tabular}{ccccccccc}
    \toprule
    & \multicolumn{8}{c}{Num tokens / bits} \\
                & 256 & 512 & 1024 & 2048 & 4096 & 8192 & 16384 &  32768 \\
    Train steps &   8 &   9 &   10 &   11 &   12 &  13  &    14 &     15 \\
    \midrule
    200k & 56.7 & 58.3 & 59.7 & 60.3 & 60.7 & 61.2 & 61.4 & 61.7 \\
    900k & 58.6 & 61.0 & 61.9 & 63.3 & 63.8 & 63.9 & 64.3 & 64.5 \\
    \bottomrule
  \end{tabular}
\end{table}

Table~\ref{tb:best_lbr} shows performance of the configuration which
obtained the best accuracy mapping VQ-VAE tokens to phonemes on
LibriSpeech.
Recognition accuracy is given at two time points: after 200k gradient descent steps, when the relative
performance of models can be assessed, and after 900k steps when the
models have converged. We did not observe overfitting with longer
training times.
Predicting the most frequent silence phoneme for all frames set an accuracy lower bound at 16\%.
A model discriminatively trained on the full 460 hour training set
to predict phonemes with the same architecture as the 25~Hz encoder
achieved 80\% framewise phoneme recognition accuracy, while a model with no
time-reduction layers set the upper bound at 88\%.

Table~\ref{tb:best_lbr} indicates that the mapping accuracy improves
with the number of tokens, with the best model reaching $64.5\%$
accuracy using 32768 tokens. 
However, the largest accuracy gain occurs at 4096 tokens, with diminishing returns as the 
number of tokens is further increased. This result is in rough correspondence with the 5760 tied triphone states used
in the Kaldi \texttt{tri6b} model.

We also note that increasing the number of tokens does not trivially
lead to improved accuracies, because we measure generalization, and not cluster purity.
In the limit of assigning a different
token to each frame, the accuracy will be poor
because of
overfitting to the small development set on which we establish the
mapping. However, in our experiments we consistently observed improved
accuracy.

\subsection{Unsupervised ZeroSpeech 2017 acoustic unit discovery}\label{sec:exp_zs}

The ZeroSpeech 2017 phonetic unit discovery task \citep{dunbar2017zero} evaluates a representation's ability to
discriminate between different sounds, rather than the ease
of mapping the representation to predefined phonetic units. It is therefore complementary to the
phoneme classification accuracy metric used in the previous section.
The ZeroSpeech evaluation scheme uses the minimal pair ABX
test \citep{schatz2013evaluating,schatz2014evaluating} which assesses the model's
ability to discriminate between pairs of three phoneme long
segments of speech that differ only in the middle phone (e.g.\ ``get''  and ``got'').
We trained the models on the provided training data (45 hours for English, 24 hours for
French and 2.5 hours for Mandarin) and evaluated them on the test data using the
official evaluation scripts. To ensure that we do not overfit to the ZeroSpeech task we
only considered the best hyperparameter settings found on
LibriSpeech\footnote{The comparison with other systems
  from the challenge is fair, because according to the ZeroSpeech
  experimental protocol, all participants were encouraged to tune their
  systems on the three languages that we use (English, French, and
  Mandarin), while the final evaluation used two surprise languages
  for which we do not have the labels required for evaluation.} (c.f.\ Section~\ref{sec:exp_hyper}). Moreover, to maximally
abide by the ZeroSpeech convention, we used the same
hyperparameters for all languages, denoted as \emph{VQ-VAE (per lang, MFCC,
\probe{cond})} in Table~\ref{tb:best_zs}.

\begin{table*}[t]
  \caption[ZeroSpeech 2017 results]{ZeroSpeech 2017 phonetic unit discovery ABX scores reported
    across- and within-speakers (lower is better).
    The VQ-VAE encoder is
    speaker independent and thus its results do not change with the
    amount of test speaker data (1s, 10s, or 2m), while
    speaker-adaptive models (e.g.\ supervised topline)
    improve with more target speaker data.  We report
    the two reference points from the challenge, along with the
    challenge winner \citep{heck2016unsupervised} and three other
    submissions that used neural network in an unsupervised setting
    \citep{chen2017multilingual,ansari2017deep,yuan2017extracting}.  
    All VQ-VAE models use
    exactly the same hyperparameter setup (14 bit tokens extracted at
    50~Hz with time-jitter probability 0.5), regardless of the amount
    of unlabeled training data (45h, 24h or 2.4h).\protect\linebreak
    The top VQ-VAE results row (VQ-VAE trained on target language,
    features extracted at the \probe{cond} point) gives best
    results overall. We also include \emph{in italics} results for
    different probe points and for VQ-VAEs jointly trained on all
    languages. Multilingual training helps Mandarin.  We also
    observe that the quantization mostly discards speaker and context
    influence. The context is however recovered in the conditioning
    signal which combines information from latent vectors at neighboring timesteps.}
  \label{tb:best_zs}
  \centering
  \setlength{\tabcolsep}{0.87ex}
  \begin{tabular}{l @{\hspace{0.7em}} rrr c rrr c rrr
                    c rrr c rrr c rrr}
    \toprule
    & \multicolumn{11}{c}{Within-speaker}
    & \multicolumn{11}{c}{Across-speaker} \\
    \cmidrule{2-12} \cmidrule{14-24}
    & \multicolumn{3}{c}{English (45h)} && \multicolumn{3}{c}{French (24h)} && \multicolumn{3}{c}{Mandarin (2.4h)}
    && \multicolumn{3}{c}{English (45h)} && \multicolumn{3}{c}{French (24h)} && \multicolumn{3}{c}{Mandarin (2.4h)} \\
    Model
    & 1s        & 10s       & 2m     && 1s        & 10s       & 2m     && 1s        & 10s       & 2m
    && 1s        & 10s       & 2m     && 1s        & 10s       & 2m     && 1s        & 10s       & 2m     \\
    \cmidrule(r){1-1}
    \cmidrule{2-4} \cmidrule{6-8} \cmidrule{10-12}
    \cmidrule{14-16} \cmidrule{18-20} \cmidrule{22-24}
    Unsupervised baseline & 12.0 & 12.1 & 12.1 && 12.5 & 12.6 & 12.6 && 11.5 & 11.5 & 11.5
    && 23.4 & 23.4 & 23.4 && 25.2 & 25.5 & 25.2 && 21.3 & 21.3 & 21.3 \\
    Supervised topline    & 6.5  & 5.3  & 5.1  && 8.0  &  6.8  & 6.8  && 9.5  & 4.2  & 4.0
    && 8.6  & 6.9  & 6.7  && 10.6  & 9.1  & 8.9  && 12.0 & 5.7  & 5.1  \\
    \addlinespace
    VQ-VAE (per lang, MFCC, \probe{cond})  &
    \textbf{5.6}  & \textbf{5.5} & \textbf{5.5}  &&
    \textbf{7.3}  & \textbf{7.5}  & \textbf{7.5}  &&
    11.2 & 10.7 & 10.8   &&
    \textbf{8.1}  & \textbf{8.0} & \textbf{8.0}  &&
    \textbf{11.0}  & \textbf{10.8}  & \textbf{11.1}  &&
    12.2 & 11.7 & 11.9   \\
    VQ-VAE (per lang, MFCC, \probe{bn})  &
    \emph{6.2} & \emph{6.0} & \emph{6.0} &&
    \emph{7.5} & \emph{7.3} & \emph{7.6} &&
    \emph{10.8} & \emph{10.5} & \emph{10.6}  &&
    \emph{8.9}  & \emph{8.8}  & \emph{8.9}  &&
    \emph{11.3} & \emph{11.0} & \emph{11.2}  &&
    \emph{11.9} & \emph{11.4} & \emph{11.6}  \\
    VQ-VAE (per lang, MFCC, \probe{proj})  &
    \emph{5.9}  & \emph{5.8}  & \emph{5.9}  &&
    \emph{6.7}  & \emph{6.9}  & \emph{6.9}  &&
    \emph{9.9}  & \emph{9.7}  & \emph{9.7}  &&
    \emph{9.1}   & \emph{9.0}   & \emph{9.0}  &&
    \emph{11.9}  & \emph{11.6}  & \emph{11.7}  &&
    \emph{11.0}  & \emph{10.6}  & \emph{10.7}   \\
    \addlinespace
    VQ-VAE (all lang, MFCC, \probe{cond}) &
    \emph{5.8}  & \emph{5.8}  & \emph{5.8}  &&
    \emph{8.0}  & \emph{7.9}  & \emph{7.8}  &&
    \emph{9.2}  & \emph{9.1}  & \emph{9.2}  &&
    \emph{8.8}   & \emph{8.6}   & \emph{8.7}  &&
    \emph{11.8}  & \emph{11.6}  & \emph{11.6}  &&
    \emph{10.3}  & \emph{10.0}  & \emph{9.9}  \\
    VQ-VAE (all lang, MFCC, \probe{bn}) &
    \emph{6.3}  & \emph{6.2}  & \emph{6.3}  &&
    \emph{8.0}  & \emph{8.0}  & \emph{7.9}  &&
    \emph{9.0}  & \emph{8.9}  & \emph{9.1}  &&
    \emph{9.4}   & \emph{9.2}   & \emph{9.3}  &&
    \emph{11.8}  & \emph{11.7}  & \emph{11.8}  &&
    \emph{9.9}   & \emph{9.7}   & \emph{9.7}  \\
    VQ-VAE (all lang, MFCC, \probe{proj}) &
    \emph{5.8}  & \emph{5.7} & \emph{5.8}  &&
    \emph{7.1}  & \emph{7.0} & \emph{6.9}  &&
    \emph{7.4}  & \emph{7.2} & \emph{7.1}  &&
    \emph{9.3}   & \emph{9.3}   & \emph{9.3}  &&
    \emph{11.9}  & \emph{11.4}  & \emph{11.6}  &&
    \emph{8.6}   & \emph{8.5}   & \emph{8.5}  \\
    VQ-VAE (all lang, fbank, \probe{proj}) &
    \emph{6.0}  & \emph{6.0} & \emph{6.0}  &&
    \emph{6.9}  & \emph{6.8} & \emph{6.8}  &&
    \emph{6.8}  & \emph{6.6} & \emph{6.6}  &&
    \emph{10.1}   & \emph{10.1}   & \emph{10.1}  &&
    \emph{12.5}  & \emph{12.2}  & \emph{12.3}  &&
    \emph{7.8}   & \emph{7.7}   & \emph{7.7}  \\

    \addlinespace
    Heck et al. \citep{heck2016unsupervised}
                          & 6.9  & 6.2  & 6.0  && 9.7  &	8.7  & 8.4  && \textbf{8.8}  & \textbf{7.9}  & \textbf{7.8}
                          && 10.1 & 8.7  & 8.5  && 13.6  & 11.7 & 11.3 && \textbf{8.8}  & \textbf{7.4}  & \textbf{7.3}  \\
    Chen et al. \citep{chen2017multilingual}
                          & 8.5	 & 7.3	& 7.2  && 11.2 & 9.4  & 9.4  && 10.5 & 8.7  & 8.5
                          && 12.7 & 11.0 & 10.8 && 17.0  & 14.5 & 14.1 && 11.9 & 10.3 & 10.1 \\
    Ansari et al. \citep{ansari2017deep}
                          & 7.7	 & 6.8	& N/A  && 10.4 & N/A  & 8.8  && 10.4 & 9.3  & 9.1
                          && 13.2 & 12.0 & N/A  && 17.2  & N/A  & 15.4 && 13.0 & 12.2 & 12.3 \\
    Yuan et al. \citep{yuan2017extracting}
                          & 9.0	 & 7.1  & 7.0  && 11.9 & 9.5  & 9.5  && 11.1 & 8.5  & 8.2
                          && 14.0 & 11.9 & 11.7 && 18.6  & 15.5 & 14.9 && 12.7 & 10.8 & 10.7 \\
    \bottomrule
  \end{tabular}
\end{table*}

On English and French, which come with sufficiently large
training datasets, we achieve results better than the
top contestant \citet{heck2016unsupervised}, despite using a
speaker independent encoder. 

The results are consistent with our analysis of information
separation performed by the VQ-VAE bottleneck: in the
more challenging across-speaker evaluation, the best performance uses
the \probe{cond} representation, which combines several neighboring frames of the bottleneck representation
(\emph{VQ-VAE, (per lang, MFCC,
  \probe{cond})} in Table~\ref{tb:best_zs}).
Comparing within- and across-speaker results is similarly consistent with the
observations in Section~\ref{sec:exp_disentag}. In the
within-speaker case, it is not necessary to disentangle speaker identity from phonetic content
so the quantization between \probe{proj} and
\probe{bn} probe points hurts performance (although on English
this is corrected by considering the broader context at \probe{cond}). In the
across-speaker case, quantization improves the scores on English and
French because the gain from discarding the confounding speaker information
offsets the loss of some phonetic details. Moreover, the
discarded phonetic information can be recovered by mixing %
neighboring timesteps at \probe{cond}.

VQ-VAE performance on Mandarin is worse, which we can attribute to three
main causes.  First, the training dataset consists of only 2.4 hours or speech, leading to
overfitting (see Sec.~\ref{sec:datasize}).  This can be partially
improved by multilingual training, as in \emph{VQ-VAE, (all lang, MFCC,
  \probe{cond})}. Second, Mandarin is a
tonal language, while the default input features (MFCCs) discard
pitch information. We note a slight improvement with a multilingual
model trained on mel filterbank features (\emph{VQ-VAE, (all lang, fbank,
  \probe{proj})}). Third, VQ-VAE was shown not to encode
prosody in the latent representation \citep{oord2017neural}. Comparing
the results across probe points, we see that Mandarin is the only
language for which the VQ bottleneck discards information and decreases
performance in the across-speaker testing regime. Nevertheless,
the multilingual prequantized features yield accuracies comparable to
\cite{heck2016unsupervised}.

We do not consider the need for more unsupervised training data to be a problem.
Unlabeled data is abundant. We believe that a more powerful model that requires and can
make better use of large amounts of unlabeled training data is
preferable to a simpler model whose performance saturates on small datasets.
However, it remains to be verified if increasing the amount
of training data would help the Mandarin VQ-VAE learn to discard less tonal
information (the multilingual model might have learned to do this %
to accommodate French and English).

\subsection{Hyperparameter impact}\label{sec:exp_hyper}
All VQ-VAE autoencoder hyperparameters were tuned on the
LibriSpeech task using several grid-searches,
optimizing for the highest phoneme recognition accuracy. 
We also validated these design choices on the English part of the ZeroSpeech challenge task.
Indeed, we found that the proposed time-jitter regularization
improved ZeroSpeech ABX scores for all 
input representations. %
Using MFCC or filterbank
features yields better scores that using waveforms, and the model
consistently obtains better scores when more tokens are used.

\subsubsection{Time-jitter regularization}\label{sec:jitter_eval}

In Table~\ref{tb:jitter} we analyze the effectiveness of the time-jitter regularization
on VQ-VAE encodings and compare it to two variants of dropout: regular dropout applied to individual dimensions %
of the encoding and dropout applied randomly to the full encoding at individual time steps. 
Regular dropout does not force the model to separate information in
neighboring timesteps.  Step-wise dropout promotes
encodings which are independent across timesteps and performs slightly
worse than the time-jitter\footnote{
The token copy probability of $0.12$ keeps a given token with probability $0.88^2=0.77$
which roughly corresponds to a $0.23$ per-timestep dropout rate}.

The proposed time-jitter
regularization greatly improves token mapping accuracy
and extends the range of
token frame rates which perform well to include 50~Hz. 
While the LibriSpeech token accuracies are comparable at 25~Hz and 50~Hz,
higher token emission frequencies are important for the ZeroSpeech AUD task, on which
the 50~Hz model was noticeably better. This behavior is due to the fact
that the 25~Hz model is prone to omitting short phones
(Sec.~\ref{sec:bitrate}), which impacts
the ABX results on the ZeroSpeech task.

\begin{figure}[t]
  \centering
  \includegraphics[width=\columnwidth, trim=0 2ex 0 0, clip]{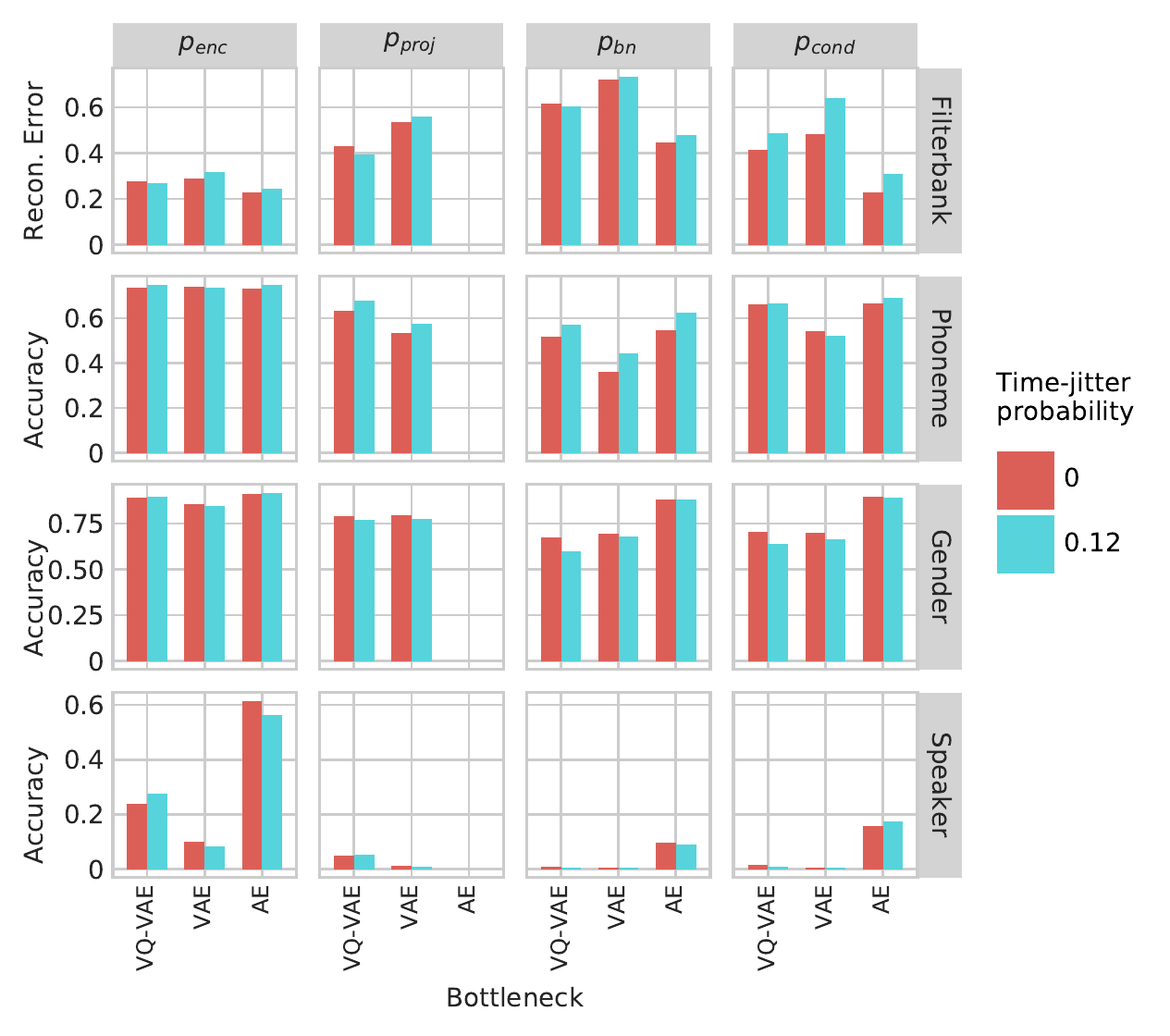}
  \caption{Impact of the time-jitter regularization on information captured by representations at different probe points.}
  \label{fig:jitter_probes}
\end{figure}

We also analyzed information content at the four probe points
for VQ-VAE, VAE, and simple dimensionality reduction AE bottleneck, shown in
Figure~\ref{fig:jitter_probes}. For all bottleneck mechanisms, the
regularization limits the quality of filterbank reconstructions and
increases the phoneme recognition accuracy in the constrained
representation. However this benefit is smaller after neighboring
timesteps are combined in the \probe{cond} probe
point. Moreover, for VQ-VAE and VAE the regularization decreases
gender prediction accuracy and 
makes the representation slightly less speaker-sensitive.

\begin{table}[t]
  \caption{Effects of input representation and regularization on %
    phoneme recognition accuracy on LibriSpeech, measured after 200k training steps.
    All models extract 256 tokens.}
  \label{tb:jitter}
  \centering
  \setlength{\tabcolsep}{0.75ex}
  \begin{tabular}{lcl@{\hspace{0.5ex}}c}
    \toprule
    Input features & Token rate & Regularization & Accuracy \\
    \midrule
    MFCC & 25 Hz & None & 52.5 \\

    MFCC & 25 Hz & Regular dropout $p=0.1$ & 50.7 \\
    MFCC & 25 Hz & Regular dropout $p=0.2$ & 49.1 \\

    MFCC & 25 Hz & Per-time step dropout $p=0.2$ & 55.3 \\
    MFCC & 25 Hz & Per-time step dropout $p=0.3$ & 55.7 \\
    MFCC & 25 Hz & Per-time step dropout $p=0.4$ & 55.1 \\

    MFCC & 25 Hz & Time-jitter  $p=0.08$ & \textbf{56.2} \\
    MFCC & 25 Hz & Time-jitter  $p=0.12$ & \textbf{56.2} \\
    MFCC & 25 Hz & Time-jitter  $p=0.16$ & 56.1 \\
    \midrule
    MFCC & 50 Hz & None & 46.5 \\
    MFCC & 50 Hz & Time-jitter $p=0.5$ & 56.1 \\
    \midrule
    log-mel spectrogram & 25 Hz & None & 50.1 \\
    log-mel spectrogram & 25 Hz & Time-jitter $p=0.12$ & 53.6 \\
    \midrule
    raw waveform & 30 Hz & None & 37.6 \\
    raw waveform & 30 Hz & Time-jitter $p=0.12$ & 48.1 \\
    \bottomrule
  \end{tabular}
\end{table}

\subsubsection{Input representation}
In this set of experiments we compared performance using different input representation: raw
waveforms, log-mel spectrograms, or MFCCs. The raw waveform encoder used 9
strided convolutional layers, which resulted in token extraction
frequency of 30~Hz.
We then replaced the waveform with a customary ASR
data pipeline: 80 log-mel filterbank features extracted every 10ms from
25ms-long windows and 13 MFCC features extracted from the
mel-filterbank output, both augmented with their first and second temporal
derivatives.  Using two strided convolution layers in the encoder led to
a 25~Hz token rate for these models.

The results are
reported in the bottom of Table~\ref{tb:jitter}. High-level features, especially MFCCs,
perform better than waveforms, because by design they discard information about pitch and provide
a degree of speaker invariance.  Using such a reduced representation forces the encoder to transmit
less information to the decoder, acting as an inductive bias toward a more speaker invariant latent encoding.

\subsubsection{Output representation}
We constructed an autoregressive decoder network that reconstructed
filterbank features rather than raw waveform samples. Inspired by recent progress in
text-to-speech systems, we implemented a Tacotron~2-like decoder
\citep{shen2017natural} with a built-in information bottleneck
on the autoregressive information flow, which was found to be critical in TTS
applications. Similarly to Tacotron~2 the filterbank features were first
processed by a small ``pre-net'', we applied generous amounts of dropout and
configured the decoder
to predict up to 4 frames in parallel.
However, these modifications yielded at best 42\% phoneme recognition
accuracy, significantly lower than the other architectures described in this paper.
The model was however an order of magnitude faster to train.

\begin{figure}[t]
  \centering
  \includegraphics[width=1.0\columnwidth, trim=0 2ex 0 0, clip]{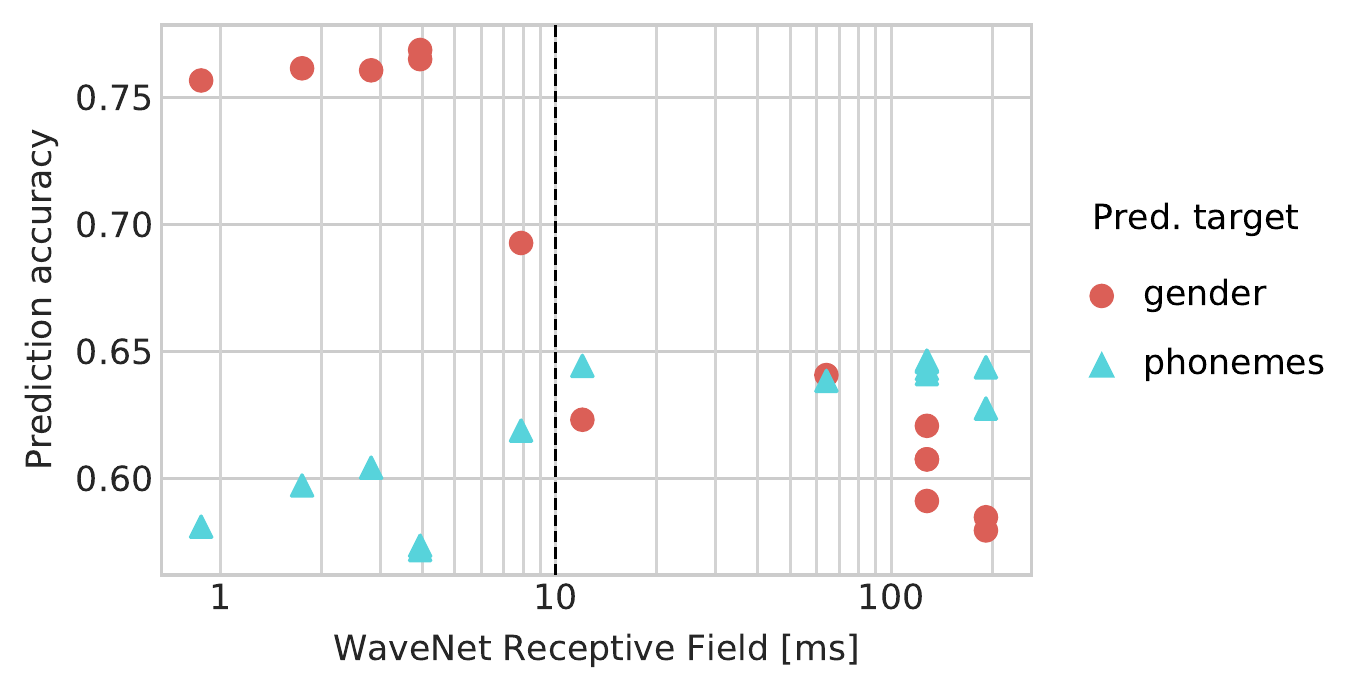}
  \caption{Impact of decoder WaveNet receptive field on the properties of the
    VQ-VAE conditioning signal. The representation is significantly more
    gender invariant when the receptive field is larger that 10ms. Frame-wise
    phoneme recognition accuracy peaks at about 125ms. The depth and width
    of the WaveNet have a secondary effect (cf.\ points with the same RF).}
  \label{fig:dec_wavenet_size}
\end{figure}

Finally, we analyzed the impact of the size of the decoding WaveNet on
the representation extracted by the VQ-VAE. We have found that overall
receptive field (RF) has a larger impact than the depth or width of the
WaveNet. In particular, a large change in the properties of the latent
representation happens when the decoder's receptive field crosses
than about 10ms. As shown in
  Figure~\ref{fig:dec_wavenet_size}, 
for smaller RFs, the conditioning signal contains more speaker
information: gender prediction is close to 80\%, while
framewise phoneme prediction accuracy is only 55\%. For larger RFs,
gender prediction accuracy is about 60\%, while
phoneme prediction peaks near 65\%. Finally, while the
reconstruction log-likelihood improved with WaveNet depth up to 30 layers, the
phoneme recognition accuracy plateaued with 20 layers. Since the WaveNet has
the largest computational cost we decided to keep the 20 layer
configuration.

\subsubsection{Decoder speaker conditioning}

The WaveNet decoder generates samples based on three sources of information:
the previously emitted samples (via the autoregressive connection), global conditioning on
speaker or other information which is stationary in time, and on the time-varying representation
extracted from the encoder. We found that disabling global speaker conditioning
reduces phoneme classification accuracy by 3 percentage points. This
further corroborates our findings about disentanglement
induced by the VQ-VAE bottleneck, which biases the model to discard information that is
available in a more explicit form.
Throughout our experiments we used a speaker-independent
encoder. However, adapting the encoder to the speaker might further
improve the results. In fact, \citet{heck2016unsupervised}
demonstrates improvements on the ZeroSpeech task using a
speaker-adaptive approach.

\subsubsection{Encoder hyperparameters}

We experimented with tuning the %
number of
encoder convolutional layers, as well as the number of filters, and the filter length. In general, performance
improved with larger encoders, however we established that the encoder's
receptive field must be carefully controlled, with the best performing encoders seeing about
0.3 seconds of input signal for each generated token.

The effective receptive field
can be controlled using two mechanisms: by carefully tuning the encoder
architecture, or by designing an encoder with a wide receptive field, but
limiting the duration of signal segments seen during training to the desired
receptive field. In this way the model never learns to use its full capacity. When
the model was trained on 2.5s long segments, an encoder with receptive field of 0.3s
had framewise phoneme recognition accuracy %
of 56.5\%, while and encoder with a receptive field of 0.8s scored only
54.3\%. When trained on segments of 0.3s, both models performed similarly.

\subsubsection{Bottleneck bit rate}\label{sec:bitrate}

The speech VQ-VAE encoder can be seen as encoding a signal using a very low
bit rate. To achieve a predetermined target bit rate, one can control both the token rate
(i.e.,\ by controlling the degree of downsampling down in the encoder strided convolutions), and the number of tokens (or equivalently the number of
bits) extracted at every step. We
found that the token rate is a crucial parameter which must
be chosen carefully, with the best results after 200k training steps obtained
at 50~Hz (56.0\% phoneme recognition accuracy %
) and 25~Hz (56.3\%). Accuracy drops abruptly at higher
token rates (49.3\% at 100~Hz), while lower rates miss very
short phones (53\% accuracy at 12.5~Hz).

In contrast to the number of tokens, the dimensionality of the VQ-VAE embedding
has a secondary effect on representation quality. We found 64 to be a good setting,
with much smaller dimensions deteriorating performance for models with a small number of tokens and
higher dimensionalities negatively affecting performance for models with a large number of tokens.

For completeness, we observe that even for the model with the largest inventory of
tokens, the overall encoder bitrate is low: 14 bits at 50~Hz = 700~bps, which is on par with the lowest
bitrate of classical speech codecs \cite{wang1998800}.

\subsubsection{Training corpus size}\label{sec:datasize}

We experimented with training models on subsets of the LibriSpeech training set, varying the size from 4.6 hours
(1\%) to 460 hours (100\%).  Training on 4.6 hours of data, 
phoneme recognition %
accuracy peaked at 50.5\% at 100k steps and then deteriorated.
Training on 9 hours led to a peak accuracy of 52.5\% at 180k sets.
When the size of training set was increased
past 23 hours the phoneme recognition reached 54\% after around 900k steps.
No further improvements were found by training on the full 460 hours of
data. We did not observe any overfitting, and for best results
trained models until reaching 900k steps with no early
stopping. An interesting future area for research would be investigating methods to increase
the model capacity to make better use of larger amounts of
unlabeled data.

The influence of the size of the dataset is also visible in the
ZeroSpeech Challenge results (Table~\ref{tb:best_zs}): VQ-VAE models obtained good
performance on English (45 hours of training data) and French (24 hours), but
performed poorly on Mandarin (2.5 hours). Moreover, on English and French we
obtained the best results with models trained on monolingual data. On
Mandarin slightly better results were obtained using a model trained
jointly on data from all languages.

\section{Related Work}\label{sec:rel_work}
VAEs for sequential data were introduced in \citep{bowman2015generating}.
The model used LSTM encoder and decoder, while the latent
representation was formed from the last hidden state of the
encoder. The model proved useful for natural language processing tasks. However, it also
demonstrated the problem of latent representation collapse: when a
powerful autoregressive decoder is used simultaneously with a
penalty on the latent encoding, such as the KL prior, the VAE
has a tendency to ignore the prior and act as if it were a purely
autoregressive sequence model. 
This issue can be mitigated by changing the weight of the KL term,
and limiting the amount of information on the autoregressive path by
using word dropout \citet{bowman2015generating}.
Latent collapse can also be avoided in deterministic autoencoders,
such as \cite{engel2017neural}, which coupled a convolutional encoder
to a powerful autoregressive WaveNet decoder \citep{oord2016wavenet}
to learn a latent representation of music audio consisting of isolated
notes from a variety of instruments.

We empirically validate that conditioning the decoder on speaker
information results in encodings which are more speaker
invariant. Moyer et al. \cite{moyer_invariant_2018} give a rigorous
proof that this approach produces representations that are invariant
to the explicitly provided information and relate it to
domain-adversarial training, another technique designed to enforce
invariance to a known nuisance factor \cite{ganin_domainadversarial_2016}.

When applied to audio, the VQ-VAE uses the WaveNet decoder to free the latent
representation from modeling information that is easily recoverable form the
recent past \cite{oord2017neural}. It avoids the problem of posterior collapse by
using a discrete latent code with a uniform prior which results in a
constant KL penalty. We employ the same strategy to design the latent
representation regularizer: rather than extending the cost function
with a penalty term that can cause the latent space to collapse, we
rely on random copies of the latent variables to prevent their
co-adaptation and promote stability over time.

The randomized time-jitter regularization introduced in this paper is inspired by slow
representations of data \citet{wiskott2002slow} and by dropout, which
randomly removes during training neurons to prevent their co-adaptation
\citet{srivastava2014dropout}. It is also very similar to Zoneout
\citep{krueger2016zoneout} which relies on random time copies of
selected neurons to regularize recurrent neural networks.

Several authors have recently proposed to model sequences with VAEs
that use a hierarchy of variables. \citet{hsu2017unsupervised} explore
a hierarchical latent space which separates sequence-dependent
variables from those which are sequence-independent ones. Their model was shown to
perform speaker conversion and to improve automatic speech recognititon (ASR) performance in the
presence of domain mismatch. \citet{li2018deep} introduce a stochastic
latent variable model for sequential data which also yields
disentangled representations and allows content swapping between
generated sequences. These other approaches could possibly benefit from regularizing
the latent representation to achieve further information
disentanglement.

Acoustic unit discovery systems aim at transducing the acoustic signal
into a sequence of interpretable units akin to phones. They often
involve clustering of acoustic frames, MFCC or neural network bottleneck features,
regularized using a probabilistic prior. DP-GMM
\citep{chen2015parallel} imposes a Dirichlet Process prior over a
Gaussian Mixture Model. Extending it with an HMM temporal structure
for sub-phonetic units leads to the DP-HMM and the HDP-HMM
\citep{lee2012nonparametric,ondel2016variational,marxer2016unsupervised}. HMM-VAE
proposes the use of a deep neural network instead of a GMM
\citep{ebbers2017hidden,glarner2018full}. These approaches enforce
top-down constraints via HMM temporal smoothing and temporal
modeling. Linguistic unit discovery models detect recurring speech
patterns at a word-like level, finding commonly repeated segments with
a constrained dynamic time warping \citep{park2008unsupervised}.

In the segmental unsupervised speech recognition
framework, neural autoencoders were used to embed variable length
speech segments into a common vector space where they could be clustered
into word types \citep{kamper2017segmental}. 
\citet{chung2017learning} replace the segmental
autoencoder with a model that instead predicts a nearby speech segment
and demonstrate that the representation shares many properties with
word embeddings. Coupled with an unsupervised word segmentation algorithm 
and unsupervised mapping of word embeddings discovered on separate corpora
\citep{lample2017unsupervised} the approach yielded an ASR system trained on
unpaired speech and text data \citep{chung2018unsupervised}.

Several entries to the ZeroSpeech 2017 challenge relied on neural
networks for phonetic unit discovery. \citet{yuan2017extracting}
trains an autoencoder on pairs of speech segments found using an
unsupervised term discovery system \citep{jansen2011efficient}.
\citet{chen2017multilingual} first clustered speech frames, then
trained a neural network to predict the cluster IDs and used its
hidden representation as features. \citet{ansari2017deep} extended
this scheme with features discovered by an autoencoder trained on
MFCCs.

\section{Conclusions}

We applied sequence autoencoders to speech modeling and compared different information bottlenecks, including VAEs and VQ-VAEs.
We carefully evaluated the induced latent representation using
interpretability criteria as well as the ability to discriminate between similar speech sounds. The
comparison of bottlenecks revealed
that discrete representations obtained using  VQ-VAE
preserved the most phonetic information while also being the most speaker-invariant. The
extracted representation allowed for accurate mapping of the extracted symbols into
phonemes and obtained competitive performance on the ZeroSpeech 2017
acoustic unit discovery task. A similar combination of VQ-VAE encoder
  and WaveNet decoder by Cho et al. had the best
acoustic unit discovery performance in ZeroSpeech 2019 \cite{dunbar_zero_2019}.

We established that an information bottleneck is required for the model to learn a
representation that separates content from speaker characteristics. Furthermore, we observe that the latent collapse problem
induced by bottlenecks which are too strong can be avoided by making the
bottleneck strength a model hyperparameter,
either removing it completely (as in the VQ-VAE), or by using the free-information %
VAE
objective.

To further improve representation quality, we introduced a time-jitter
regularization scheme which limits the capacity of the latent code yet does not
result in a collapse of the latent space. We hope that this can similarly
improve performance of latent variable models used with
auto-regressive decoders in other problem domains.

Both the VAE and VQ-VAE constrain the information bandwidth of the latent
representation. However, the VQ-VAE uses a quantization mechanism,
which deterministically forces the encoding to be equal to a
prototype, while the VAE limits the amount of information by injecting noise.
In our study, the VQ-VAE resulted in better information
separation than the VAE. However, further experiments are needed to
fully understand this effect. In particular, is this a consequence of the
quantization, or of the deterministic operation? 

We also observe that while the VQ-VAE produces a discrete
representation, for best results it uses a token set so large that it
is impractical to assign a separate meaning to each one. In
particular, in our ZeroSpeech experiments we used the dense embedding representation of each token,
which provided a more nuanced token similarity measure than
simply using the token identity. Perhaps a more structured latent
representation is needed, in which a small set of units
can be modulated in a continuous fashion.

Extensive hyperparameter evaluation indicated that optimizing the receptive field sizes of the encoder and decoder networks
is important for good model performance. A multi-scale modeling approach could
furthermore separate the prosodic information. Our autoencoding approach could also be combined with penalties that are 
more specialized to speech processing. Introducing a HMM prior as in \citet{glarner2018full}
could promote a latent representation which better mimics the temporal phonetic structure
of speech.

\section*{Acknowledgments}
The authors thank
Tara Sainath, Úlfar Erlingsson, Aren Jansen,
Sander Dieleman, Jesse Engel, Łukasz Kaiser, Tom Walters, Cristina Garbacea,
and the Google Brain team for their helpful discussions and feedback.

\bibliographystyle{IEEEtran}
\bibliography{paper}

\iftrue
\begin{IEEEbiography}[{\includegraphics[width=1in,height=1.25in,clip,keepaspectratio]{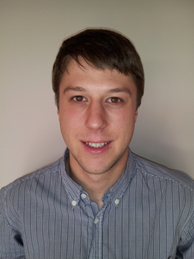}}]
  {Jan Chorowski} is an Associate Professor at Faculty of Mathematics and
  Computer Science at the University of Wrocław. He received his
  M.Sc. degree in electrical engineering from the Wrocław University of
  Technology, Poland and EE PhD from the University of Louisville,
  Kentucky in 2012. He has worked with several research teams, including
  Google Brain, Microsoft Research and Yoshua Bengio’s lab at the
  University of Montreal.  His research interests are applications of
  neural networks to problems which are intuitive for humans
  but difficult for machines, such as speech and natural language
  processing.
\end{IEEEbiography}

\begin{IEEEbiography}[{\includegraphics[width=1in,height=1.25in,clip,keepaspectratio]{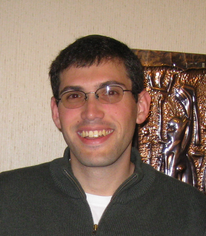}}]
  {Ron J.~Weiss} is a software engineer at Google where he has worked on
  content-based audio analysis, recommender systems for music, noise
  robust speech recognition, speech translation, and speech synthesis. Ron
  completed his Ph.D. in electrical
  engineering from Columbia University in 2009 where he worked in the
  Laboratory for the Recognition of Speech and Audio. From 2009 to 2010
  he was a postdoctoral researcher in the Music and Audio Research
  Laboratory at New York University.
\end{IEEEbiography}

\begin{IEEEbiography}[{\includegraphics[width=1in,height=1.25in,clip,keepaspectratio]{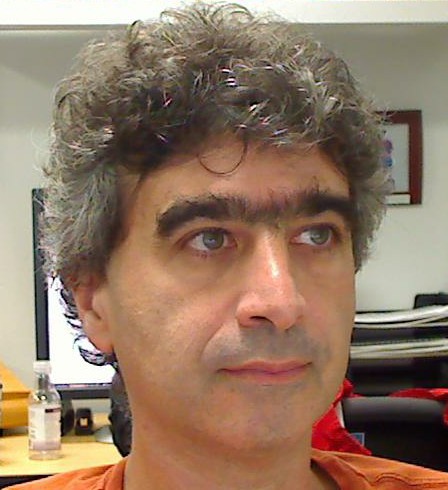}}]
{Samy Bengio} (PhD in computer science, University of Montreal, 1993) is
a research scientist at Google since 2007. He currently leads a group
of research scientists in the Google Brain team, conducting research
in many areas of machine learning such as deep architectures,
representation learning, sequence processing, speech recognition,
image understanding, large-scale problems, adversarial settings, etc.

He is the general chair for Neural Information Processing Systems
(NeurIPS) 2018, the main conference venue for machine learning, was
the program chair for NeurIPS in 2017, is action editor of the Journal
of Machine Learning Research and on the editorial board of the Machine
Learning Journal, was program chair of the International Conference on
Learning Representations (ICLR 2015, 2016), general chair of BayLearn
(2012-2015) and the Workshops on Machine Learning for Multimodal
Interactions (MLMI'2004-2006), as well as the IEEE Workshop on Neural
Networks for Signal Processing (NNSP'2002), and on the program
committee of several international conferences such as NeurIPS, ICML,
ICLR, ECML and IJCAI.
\end{IEEEbiography}

\begin{IEEEbiography}[{\includegraphics[width=1in,height=1.25in,clip,keepaspectratio]{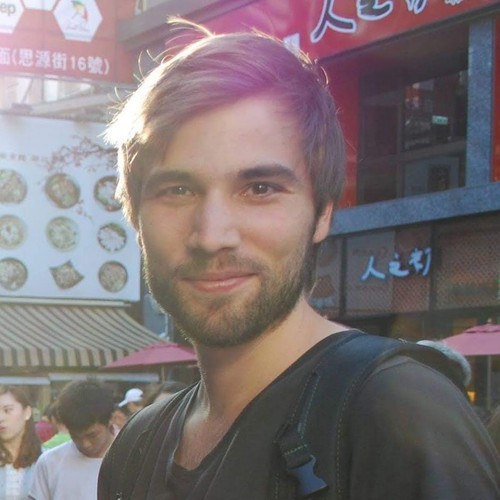}}]
  {A\"{a}ron van den Oord} is a research scientist at DeepMind,
  London.  A\"{a}ron completed his PhD at the University of Ghent,
  Belgium in 2015.  He has worked on unsupervised representation
  learning, music recommendation, generative modeling with
  autoregressive networks and various applications of generative
  models such text-to-speech synthesis and data compression.
\end{IEEEbiography}
\fi

\end{document}